\documentclass[runningheads]{llncs}
\usepackage[T1]{fontenc}

\usepackage{amssymb}
\usepackage{amsmath}

\usepackage{hyperref}

\usepackage{graphicx}
\usepackage{color}

\usepackage{subcaption}
\usepackage{caption}
\usepackage{booktabs}
\usepackage{makecell}
\usepackage{algorithm}
\usepackage{algpseudocode}
\usepackage{multirow}
\usepackage[misc]{ifsym}
\usepackage{ltablex}
\usepackage{comment}
\usepackage{todonotes}

\begin{document}
\title{One-Class Intrusion Detection with Dynamic Graphs %
\thanks{We gratefully acknowledge funding by the BMBF within the project HAIP, grant number 16KIS1212}}

%
%
\author{Aleksei Liuliakov\orcidID{0000-0003-4676-9272}
\and
Alexander Schulz\orcidID{0000-0002-0739-612X}
\and
Luca Hermes\orcidID{0000-0002-7568-7981}
\and
Barbara Hammer\orcidID{0000-0002-0935-5591}}
\authorrunning{A. Liuliakov et al.}
%
\institute{Machine Learning Group, Bielefeld University, Germany
\email{\{aliuliakov|aschulz|lhermes|bhammer\}@techfak.uni-bielefeld.de}}
\maketitle              
\begin{abstract}
With the growing digitalization all over the globe, the relevance of network security becomes increasingly important. Machine learning-based intrusion detection constitutes a promising approach for improving security, but it bears several challenges. These include the requirement to detect novel and unseen network events, as well as specific data properties, such as events over time together with the inherent graph structure of network communication. 

In this work, we propose a novel intrusion detection method, \emph{TGN-SVDD}, which builds upon modern dynamic graph modelling and deep anomaly detection. We demonstrate its superiority over several baselines for realistic intrusion detection data and suggest a more challenging variant of the latter. Our implementation is available online.\footnote{\url{https://github.com/AlekseiLiu/tgn_svdd}}

\keywords{Temporal dynamic graph \and One class classification \and Intrusion detection.}
\end{abstract}
\section{Introduction}

The field of anomaly detection deals with detecting rare observations, sometimes also referred to as outliers or novelties, that differ substantially from the majority of samples. This is approached (mostly) in a fully unsupervised fashion with only regular samples being available. The interest in this problem has been increasing in recent years due to a growing potential impact in different areas, such as security, medicine or finance. A variety of successful models for this problem has been proposed, ranging from shallow approaches like the One-Class Support Vector Machine (OCSVM), the Support Vector Data Description (SVDD), Isolation Forest (IF) or Local Outlier Factor (LOF) \cite{scholkopf2001estimating,tax2004svdd,liu2008isolation,breunig2000lof}, over deep methods like Deep SVDD or Deep OCSVM \cite{ruff2018deep,erfani2016high}, to graph based ones like the  Temporal Hierarchical One-Class (THOC) network or Event2Graph \cite{shen2020timeseries,wu2021event2graph}. Several more have been discussed in survey articles focusing on specific aspects, such as deep or graph based models \cite{ruff2021unifying,kwon2019survey,pang2021deep}. This field has been investigated from different areas, including automated machine learning based approaches \cite{liuliakov2023automl}.

In the present work we want to focus on the subfield of intrusion detection in computer networks. Specifically, we aim to detect abnormal network traffic that constitutes an attack by an intruder. This is a relevant topic, because the size and abundance of computer networks keeps increasing. Accordingly, the dependence of the public and the private sector on the former is ever-growing. This makes the potential danger and costs of attacks on such networks, such as network intrusion attacks, evident. In this domain, specific properties of the data are present that are not necessarily typical for classical anomaly detection: First, network communication appears sequentially over time, making dynamical data structures promising candidates; second, communication has the structure of a sender, a recipient and a communication message \cite{Sharafaldin2018TowardGA}, which can be most naturally represented as graphs. For this purpose, Dynamic Graph Neural Networks are a promising model class, including the approaches \cite{rossi2020temporal,kumar2019predicting,trivedi2019dyrep,xu2020inductive,wang2021inductive}. In a recent study \cite{poursafaei2022better}, the Temporal Graph Network (TGN) \cite{rossi2020temporal}, has shown to be particularly successful in modelling dynamical network data. However, empirically, the TGN is not sufficient for intrusion detection. Hence, we propose an extension of this model and evaluate it in the context of intrusion detection.

Our contributions are the following: 

\begin{itemize}
  \item We propose a new fully unsupervised end-to-end trainable intrusion detection model, that we call \emph{TGN-SVDD}, which utilizes dynamic graph modelling and combines the two approaches TGN and Deep SVDD.
  \item We demonstrate that the vanilla TGN is not sufficient for intrusion detection in realistic benchmark data \cite{Sharafaldin2018TowardGA}, while performing better than shallow models. Both are outperformed by our proposed TGN-SVDD.
  \item We analyze the dataset \cite{Sharafaldin2018TowardGA} in depth and detect a potential easy workaround that could be used by models for intrusion detection. We suggest a solution to this problem making the dataset more challenging and show that our proposed method still achieves a high performance level.
\end{itemize}

\section{Fundamentals}

Network traffic and more specifically internet traffic refers to the collective flow of data packets transmitted, received, and routed between interconnected devices and systems. This traffic encompasses various data types, including text, multimedia, and control information exchanged through a diverse set of application-layer protocols such as HTTP, FTP, SMTP, etc. Internet traffic can be represented as a set of network flows, where each is a sequence of data packets that share common attributes, such as the source IP address, destination IP address, source port, destination port, and protocol. It can be further conceptualized as dynamic temporal graphs. In this representation, source and destination are represented as nodes and identified by their respective IP address and the connections between these nodes, defined by network flows, act as the edges or links. Note that each edge is associated with a timestamp that reflects when that particular network flow appeared.

\subsection{Continuous-Time Dynamic Graphs (CTDG)}

Continuous-time dynamic graphs (CTDG) are represented as timed event lists, including edge or node addition/deletion and feature transformations. Temporal (multi-)graphs are sequences of time-stamped events $G = \{x(t_1), x(t_2), ...\}$, with events $x(t)$ adding/changing nodes or interactions. There are two event types: 1) node-wise events $\mathbf{v}_i(t)$ such as creating a node $i$ or updating its features, and 2) interaction events in the form of directed temporal edges $\mathbf{e}_{ij}(t)$.

Denote $V(T) = \{i : \exists\, \mathbf{v}_i(t) \in G, t \in T\}$ and $E(T) = \{(i, j) : \exists\, \mathbf{e}_{ij}(t) \in G, t \in T\}$ as temporal vertex and edge sets, and $\mathcal{N}_i(T) = \{j : (i, j) \in E(T)\}$ as node $i$'s neighborhood in time interval $T$. $\mathcal{N}_{i}^{\nu}(T)$ is the $\nu$-hop neighborhood. A snapshot of graph $G$ at time $t$ is the (multi-)graph $G(t) = \left(V([0, t]), E([0, t])\right)$.

\subsection{Temporal Graph Network (TGN)}
\label{sec:tgn}
One popular framework to model dynamic graphs is the TGN \cite{rossi2020temporal}. TGN model with graph attention mechanism consists of an encoder-decoder pair for dynamic graph analysis. The TGN encoder for continuous-time dynamic graphs generates node embeddings that capture long-term dependencies. The decoder uses the embeddings to make task-specific predictions.

The model maintains a vector for each node as memory which represents the compressed history. Messages are computed for each node participating in the event. Separate message functions for source, target, and node-wise changes are used.
After each event node memory is updated by means of a learnable memory function (e.g.\ GRU) with messages and previous memory states as inputs respectively. This enables the model to capture long-term dependencies.

In a given interaction event between nodes $i$ and $j$ at time step $t$, a Temporal Graph Attention Module effectively incorporates the historical events of either node. For node $i$, the module retrieves its current representation along with its previous interactions with other nodes. These past interactions are then weighted according to the attention mechanism and subsequently aggregated to provide a representation of the node's temporal dynamics. The output of this module results in an embedding vector for a particular node i in time t.

To define the model we use the same notation as above for CTDG. $G = \{x(t_1), ..., x(t_D)\}$ is a sequences of time-stamped and time-ordered interaction events $x(t)$, where $D$ is the number of events in the data. For every $t_k$ with $k \in \{1, .., D\}$, we have certain source and destination nodes pairs $(i, j) \in E(t_k)$, and corresponding event feature vectors $\mathbf{e}_{ij}(t_k)$. We denote a TGN memory state $\mathbf{s}_i(t_{k}^-)$ of the node $i$ at the time $t_{k}^-<t_k$, which gets updated every time when node $i$ appears in an event. We denote TGN encoder functional module $\mathbf{z}(i, \mathbf{s}_i(t_{k}^-), \mathcal{N}_i(t_k), \mathbf{W})$, which provides a vector representation of the node $i$ in embedding space $\mathcal{F} \subseteq \mathbb{R}^{p}$ with respect to the past temporal events at time $t_{k}^-$, the history of this node $s_i(t_{k}^-)$ and with respect to its temporal neighborhood $\mathcal{N}_i(t_k)$.  $\mathbf{W}$ are TGN's encoder model parameters.

In the original work, a multi-layer perceptron (MLP) decoder is employed by the authors for self-supervised next edge (event) prediction task. In our work we only utilize TGN's encoder part to obtain node embeddings that we complement with a decoder specialized for one-class classification (s. Sec. \ref{sec:tgn_svdd}).

\subsection{Deep Support Vector Data Description (Deep SVDD)}

One-Class classification focuses on learning the target class representation to identify novel or outlier instances. Traditional shallow methods like OCSVM and SVDD \cite{scholkopf2001estimating,tax2004svdd} face scalability issues and struggle in complex high-dimensional scenarios. Deep SVDD \cite{ruff2018deep} addresses these limitations by learning a feature space representation in an end-to-end setting with a deep network. It also improves performance and scalability in one-class classification and anomaly detection tasks. Deep SVDD can be integrated with various deep learning encoder architectures, leveraging recent successes in deep representation learning.

For a given input space $\mathcal{X} \subseteq \mathbb{R}^{d}$ and output space $\mathcal{F} \subseteq \mathbb{R}^{p}$, let $\phi(\cdot ; \mathbf{W}) : \mathcal{X} \rightarrow \mathcal{F}$ represent a deep neural network with parameters $\mathbf{W}$.
For any test point $\mathbf{x} \in \mathcal{X}$, an anomaly score $\mathbf{s}$ is defined by calculating the squared distance between the point and the center of a hypersphere $\mathbf{c}$. This can be expressed as:

\begin{equation}\label{eq:deep_svdd_score}
\begin{aligned}
s(\mathbf{x}) = \lVert\, \phi(\mathbf{x}; \mathbf{W}) - \mathbf{c} \,\rVert^2 \\
\end{aligned}
\end{equation}

The training data is represented as $\mathcal{D}_{\mu} = \{\mathbf{x}_1, \ldots, \mathbf{x}_{\mu}\}$, where $\mathbf{x}_i\in \mathcal{X}, \forall i \in \{1, \dots, \mu\}$. Where $\mu\in\mathbb{N}$ is a number of data points. The objective of Deep SVDD can be formulated as following:

\begin{equation}\label{eq:deep_svdd}
\begin{aligned}
\min_{\mathbf{W},\mathbf{c}} \quad & \frac{1}{\mu} \sum_{i=1}^{\mu} \left\lVert\, \phi(\mathbf{x}_i; \mathbf{W}) - \mathbf{c} \,\right\rVert^2 + \lambda \,\lVert \mathbf{W} \,\rVert^2 ,\\
\end{aligned}
\end{equation}

\noindent aims to minimize the sum of the squared distances between the network representations of input data points $\phi(\mathbf{x}_i; \mathbf{W})$ and the center $\mathbf{c} \in \mathcal{F}$ of the hypersphere, along with a weight decay regularizer term for model parameters $\mathbf{W}$, which is controlled by the hyperparameter $\lambda$. Note that $\mathbf{c}$ is optimized jointly with the network parameters. 

\section{Our Proposed Model: TGN-SVDD}
\label{sec:tgn_svdd}

In the application case of cybersecurity and Network Intrusion Detection Systems (NIDS) usually only normal/benign data is available. At the same time attacks exhibit a wide range of characteristics and new attack types may be found. Thus, training data cannot be assumed to cover all possible attacks. This makes standard supervised Machine Learning techniques suboptimal for such data and applications. We introduce a novel end-to-end trainable unsupervised approach which is best suited for, but not limited to, cybersecurity and NIDS applications. 

For the TGN encoder functional module, from section \ref{sec:tgn}, we will use the notation $\mathbf{z}_i(t_k, \mathbf{W})$ for brevity. The rest of the notation remains unchanged.

We apply a modified Deep SVDD decoder to compute an anomaly score for each given interaction event $x(t_k)$ as follows:

\begin{equation}\label{eq:tgn_svdd_score}
\begin{aligned}
s(\mathbf{x}(t_k)) = \lVert\, \left(\mathbf{z}_i(t_k, \mathbf{W})\oplus\mathbf{z}_j(t_k, \mathbf{W})\right) - \mathbf{c} \,\rVert^2, \\
\end{aligned}
\end{equation}

\noindent where $\mathbf{z}_i$ and $\mathbf{z}_j$ are the temporal node embeddings of nodes $i$ and $j$ that participate in event $x(t_k)$, $\oplus$ denotes concatenation, and, as in Deep SVDD, $\mathbf{c}$ is a trainable vector that points to the center of a hypersphere.
At initialization time, the node's memory states are set to zero-vectors. The end-to-end training objective is defined as

\begin{equation}\label{eq:tgn_svdd}
\begin{aligned}
\min_{\mathbf{W},\mathbf{c}}  \quad & \frac{1}{D} \sum_{k = 1}^D \left\lVert\, \left(\mathbf{z}_i(t_k, \mathbf{W})\oplus\mathbf{z}_j(t_k, \mathbf{W})\right) - \mathbf{c}\, \right\rVert^2 + \lambda \lVert \,\mathbf{W} \,\rVert^2, \\
\end{aligned}
\end{equation}

\noindent which aims to minimize the sum of the squared distances between the concatenated TGN encoder representations of source node $i$ and destination node $j$ and the center $\mathbf{c} \in \mathbb{R}^{2 \times p}$ of the hypersphere, along with a weight regularization term for TGN encoder model parameters $\mathbf{W}$, and corresponding trade-off hyperparameter $\lambda$. 

\section{Experiment}

In the following, we describe our performed experimentation, including the setup, the utilized data and pre-processing as well as the final results.

\subsection{Dataset and Experimental Setup}

\textbf{Dataset}
To evaluate our proposed model, we employed the CIC-IDS2017 dataset \cite{Sharafaldin2018TowardGA}, which was created by the University of New Brunswick. This publicly available dataset offers realistic intrusion detection scenarios for evaluation. The dataset was generated by designing two separate networks: the Victim-Network and the Attack-Network. The authors proposed a B-profile system to replicate background traffic, capturing the abstract behavior of 25 users based on HTTP, HTTPS, FTP, SSH, and email protocols for normal traffic. The attack traffic incorporates six attack profiles, including Brute Force, Heartbleed, Botnet, DoS, DDoS, Web, and Infiltration attacks. Data collection encompasses data gathering over five working days Monday to Friday, with Monday featuring only benign traffic and the other days containing various attacks.

\begin{table}[t]
\centering
  \caption{Statistics of the resulting data for the days that includes attacks.}
  \label{tab:data_des}
  \vspace{2mm}
  \setlength{\tabcolsep}{12pt}
  \begin{tabular}{ l c c c }
\toprule
Name & Events & Nodes & Features\\
\midrule
Tuesday & 572087 & 12972 & 61\\
Wednesday & 597202 & 13595 & 61\\
Thursday & 614336 & 13611 & 61\\
Friday & 753468 & 13314 & 61\\

\bottomrule
\end{tabular}
\end{table}

To format the raw PCAP files provided by the authors for compatibility with the model, we pre-processed the data. As dynamic temporal graphs require a sequence of timestamped events as input data, it is common to use a temporal adjacency list table format. This table includes columns for source node ID, destination node ID, timestamp, and a vector of features corresponding to the event. If applicable, an additional column for event labels may be included. We choose Network Flows (NetFlow) as source of the timestamped sequence of events, with source and destination IP addresses as unique node IDs.

To convert raw traffic into an adjacency list of timestamped NetFlow events, we utilised the NFStream framework \cite{AOUINI2022108719}. This allows us to extract a list of timestamped NetFlows along with 61 custom statistical 'core' and 'postmortem' features. Raw IP addresses are enumerated to unique IDs, and the timestamp is set to the first appearance of the first flow's packet. All continuous features are scaled to the [0, 1] interval. We labeled the data according to the attacker IP, victim IP, and time frame during which each attack was conducted, resulting in timestamped NetFlow event lists for each working day of the experiment.

Our model requires a strict sequential order, with normal data streams occurring earlier in the training phase and actual attacks appearing later in the testing phase. To accomplish this, we modified the data as follows. Since Monday only included normal traffic activities, we concatenated Monday's event list with one of the other working days (Tuesday, Wednesday, Thursday, or Friday) while respecting the timestamps. This resulted in four temporal dynamic graphs, each starting with Monday's events and continuing with malicious traffic from one of the subsequent working days.

We subtracted the largest timestamp from every event's timestamp in both parts of each data day-pair and added the largest timestamp from the first part (Monday) to the second part (one of the malicious days). This eliminates temporal discontinuity in the data (night gap between working days activity), and results in timestamps starting at 0 and monotonically increasing up to the end of the dataset. This modification is considered valid without significantly affecting the data pattern, as we are interested in intraday activity rather than intra -week, -month, or -year scales. We 
assume that events within days are similarly distributed over time. 
Details about the data are provided in Table \ref{tab:data_des}.

In this study, we partitioned the data into train, validation, and test subsets for each day, adhering to a consistent split criterion across all four datasets. The train subset comprises the initial 200,000 events, while the validation subset encompasses the subsequent 70,000 events. The remaining events constitute the test subset. The data splitting was conducted with respect to the timeline to ensure that the train and validation subsets contain only normal events, with all attacks appearing only in the test set.

\noindent \textbf{Experimental Setup} The proposed model was implemented in Python 3.9 using the Pytorch \cite{paszke2019pytorch} and PyG \cite{fey2019fast} packages.

The baselines LOF and IF are provided in the sckit-learn package \cite{pedregosa2011scikit}, the vanilla TGN baseline algorithm by the PyG package example implementations.

For our model implementation we use the TGN's encoder part from PyG, with the following parameters: time embedding dimension 200, memory and node embedding dimensions both 200. The remaining parameters are chosen as they were provided by the default model. TGN-SVDD was trained over 25 epochs.

The number of neighbors in LOF was 20, the remaining parameters default. For IF default parameters are employed. We ran a vanilla TGN model as an additional baseline using the default parameters provided in PyG.

\subsection{Results}
\begin{figure}[t]
\centering
\includegraphics[width=0.9\textwidth]{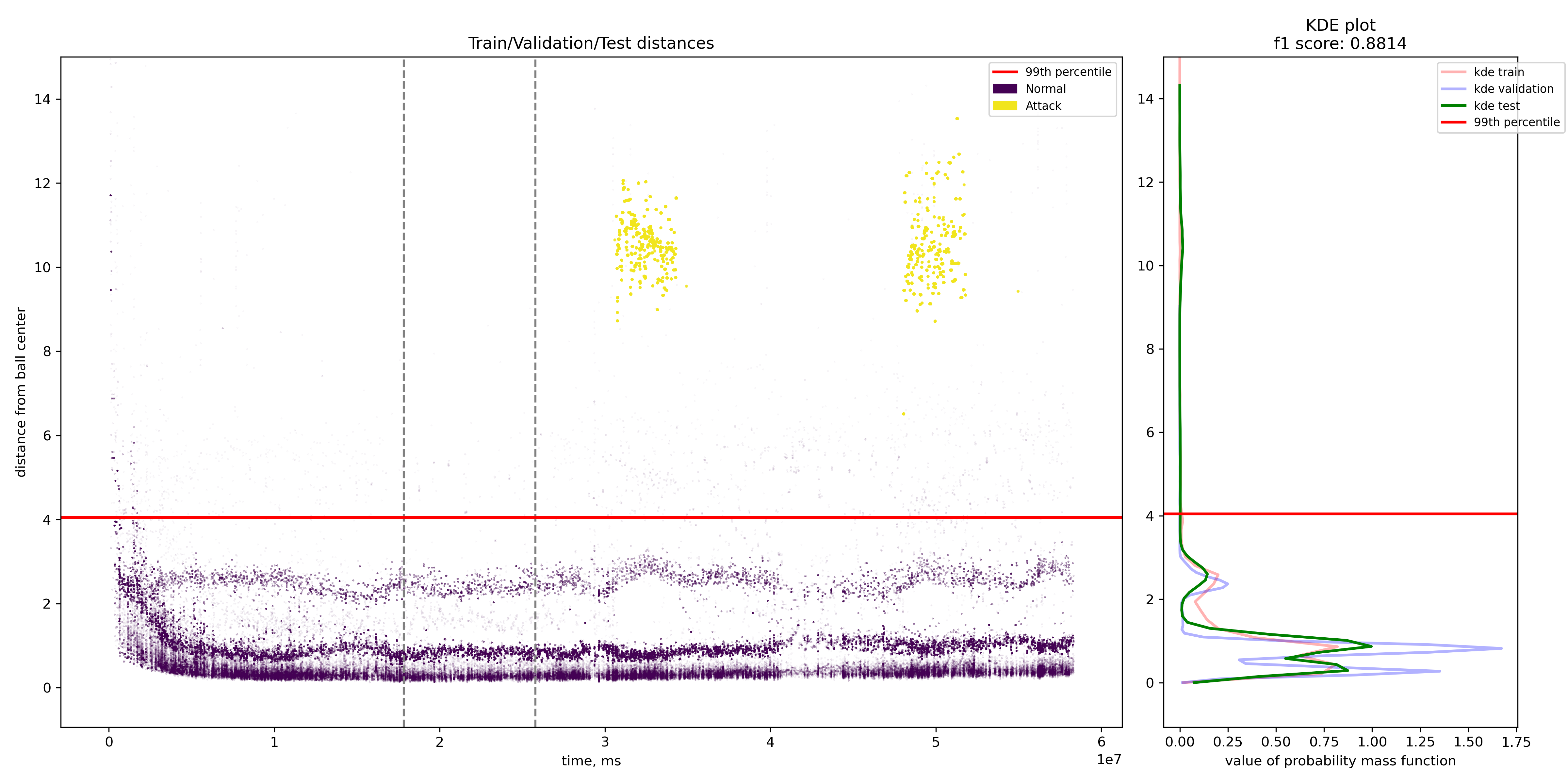}
\caption{Tuesday working hours. Illustration of TGN-SVDD performance. Left: On the y-axis, the anomaly score is depicted as it described in the model description. The two vertical lines imply the separation between training, validation and testing data. The red line shows the 99th percentile from the train set as a threshold. Right: Density estimation.} \label{fig:tue}
\end{figure}

\begin{figure}[tbp]
\centering
\includegraphics[width=0.9\textwidth]{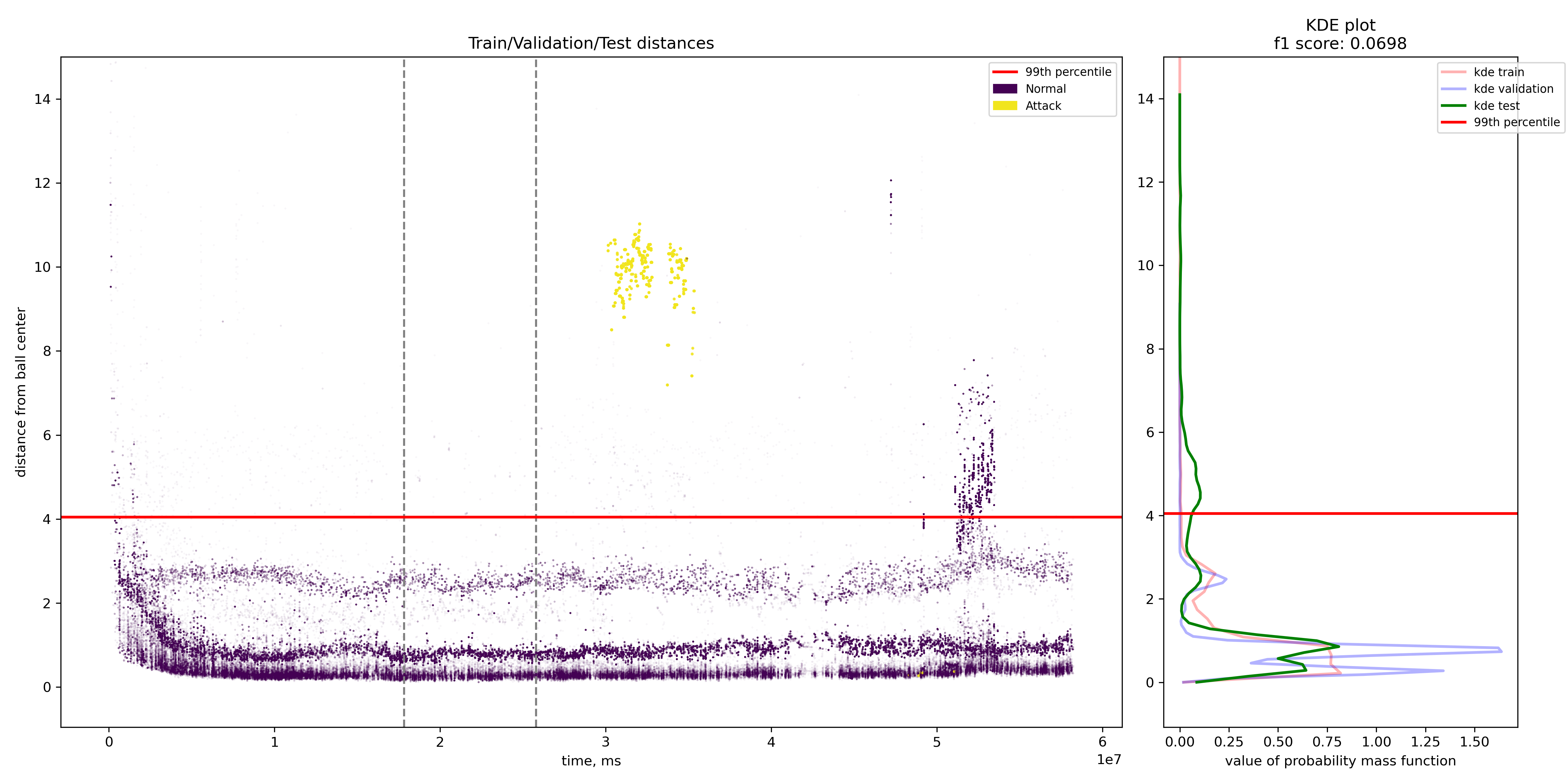}
\caption{Thursday working hours. Illustration of TGN-SVDD performance. Left: On the y-axis, the anomaly score is depicted as it described in the model description. The two vertical lines imply the separation between training, validation and testing data. The red line shows the 99th percentile from the train set as a threshold. Right: Density estimation.} \label{fig:thu}
\end{figure}

In this section, we present the evaluation results of our TGN-SVDD model and the baseline models: Vanilla TGN, LOF (novelty), LOF (outlier) and IF. The evaluation metrics, including precision, recall, F1-score and ROC AUC, are provided in the Table \ref{tab:feat_metrics_fri}; Figures \ref{fig:tue} and \ref{fig:thu} illustrate the performance of our model.

We conducted the evaluation under two different scenarios. In the first scenario, we used temporal event data \textit{with} features as input for our TGN-SVDD model and the baseline vanilla TGN model. In the second scenario, we set all event-related features to 0, which is equivalent to the case \textit{without} features at all. In this scenario TGN-SVDD and vanilla TGN rely solely on the temporal graph dynamics of the data.

The other baseline models, LOF (novelty), LOF (outlier), and Isolation Forest, were evaluated on the exact same data, including source/destination node IDs and timestamps, both with and without features. The LOF (novelty) model, as a novelty detector, was trained on the training data, and inference was performed on the testing data. LOF (outlier) and Isolation Forest, as outlier detectors, were evaluated directly on the testing data. The \textit{contamination} parameter was computed from the data as the ratio of inliers and outliers and explicitly passed to both models.

For LOF (novelty), LOF (outlier), and Isolation Forest, we used default inference settings from the scikit-learn library. For the TGN-SVDD model and the baseline vanilla TGN model, we applied a 0.99 percentile threshold obtained on the training set and used this threshold to infer labels on the test set.

ROC AUC metrics require the model to output scores for inference. For LOF (novelty), LOF (outlier), and Isolation Forest, we used local outlier factor and Isolation Forest anomaly score as measures of data point anomaly. As TGN-SVDD directly computes anomaly scores, we were able to compute ROC AUC directly. For the baseline vanilla TGN model, we chose $\text{score} = 1 - p$, where $p$ is the probability of the event to occur, meaning the higher this score, the more likely the event is an outlier.

Results are shown in the Table \ref{tab:feat_metrics_fri}. Proposed TGN-SVDD model outperforms all baseline models in both scenarios and on all datasets. In the scenario with features, Isolation Forest performed remarkably close to our method in terms of ROC AUC metric on the Wednesday dataset. LOF (novelty) for Friday showed the second-best result, significantly outperforming other baseline models. In the scenario without features, remarkable second-best ROC AUC results were shown by LOF (novelty) on Friday and Isolation Forest on Monday and Thursday.

The Attack class in the CIC-IDS2017 dataset comprises multiple specific attacks. To demonstrate the performance of our model across different Attack classes, we present the confusion matrix in Table \ref{tab:multiclass_results}. Given that TGN-SVDD is a novelty detector, it only predicts 'Normal' or 'Attack' classes for each network event. As evident from the table, the model accurately predicts the majority of Attacks, with the exceptions of 'Bot' on Friday and 'Infiltration' on Thursday.



\begin{table}[t]
\centering
  \caption{Resulting performance evaluated on several different metrics. Results are for data with event features (left) and without (right).}
  \label{tab:feat_metrics_fri}
  \vspace{1mm}
  \begin{tabular}{ l  c c c c  c c}
\toprule
  & \multicolumn{4}{c}{with features} & \multicolumn{2}{c}{without features} \\
  \cmidrule(l{.25em}){2-5} \cmidrule(l{.25em}){6-7}
  &  Precision &    Recall &  F1-score &  ROC AUC & F1-score &  ROC AUC \\ 
\cmidrule(r{.25em}){1-1} \cmidrule(l{.25em}){2-5} \cmidrule(l{.25em}){6-7}
\multicolumn{7}{c}{\textit{Tuesday}}\\

TGN-SVDD         &   0.783 &  1.000 &  0.878 &              0.999  &  0.931 &           0.999\\
LOF (novelty) & 0.023 & 1.000 & 0.045 & 0.484 & 0.045 & 0.484 \\
LOF (outlier) & 0.044 & 0.044 & 0.044 & 0.615 & 0.044 & 0.615 \\
Isolation Forest & 0.000 & 0.000 & 0.000 & 0.760 & 0.000 & 0.701 \\
TGN & 0.000 & 0.000 & 0.000 & 0.690 & 0.000 & 0.173 \\
\cmidrule(r{.25em}){1-1} \cmidrule(l{.25em}){2-5} \cmidrule(l{.25em}){6-7}
\multicolumn{7}{c}{\textit{Wednesday}} \\
TGN-SVDD         &   0.930 &  1.000 &  0.964 &            0.999 & 0.967 & 0.999\\
LOF (novelty) & 0.072 & 1.000 & 0.134 & 0.354 & 0.134 & 0.354\\
LOF (outlier) & 0.027 & 0.027 & 0.027 & 0.346 & 0.031 & 0.347\\
Isolation Forest & 0.588 & 0.588 & 0.588 & 0.946 & 0.000 & 0.167\\
TGN & 0.000 & 0.000 & 0.000 & 0.268 & 0.000 & 0.390\\
\cmidrule(r{.25em}){1-1} \cmidrule(l{.25em}){2-5} \cmidrule(l{.25em}){6-7}
\multicolumn{7}{c}{\textit{Thursday}} \\
TGN-SVDD         &   0.035 &  0.997 &  0.068 &              0.994 & 0.056 & 0.992\\
LOF (novelty) & 0.005 & 1.000 & 0.011 & 0.209 & 0.011 & 0.209\\
LOF (outlier) & 0.000 & 0.000 & 0.000 & 0.680 & 0.000 & 0.679\\
Isolation Forest & 0.006 & 0.006 & 0.006 & 0.626 & 0.000 & 0.796\\
TGN & 0.000 & 0.000 & 0.000 & 0.459 & 0.000 & 0.072\\

\cmidrule(r{.25em}){1-1} \cmidrule(l{.25em}){2-5} \cmidrule(l{.25em}){6-7}
\multicolumn{7}{c}{\textit{Friday}} \\

TGN-SVDD         &   0.992 &  0.993 &  0.993 &  0.995 & 0.991 & 0.994\\
LOF (novelty) & 0.424 & 1.000 & 0.596 & 0.813 & 0.596 & 0.813\\
LOF (outlier) & 0.291 & 0.291 & 0.291 & 0.449 & 0.244 & 0.411\\
Isolation Forest & 0.237 & 0.237 & 0.237 & 0.222 & 0.006 & 0.005\\
TGN & 0.000 & 0.000 & 0.000 & 0.613 & 0.000 & 0.295\\

\bottomrule
\end{tabular}
\end{table}

\subsection{Deeper Dive Into Data}

The dataset is structured such that the majority of malicious activities originate from a single source IP, often targeting the same destination IP - these nodes ids are 32 and 11, respectively, in our dataset's nodes enumeration. Upon further investigation, it was found that while node 11 participated in numerous normal events, node 32 was exclusively present during the testing phase, potentially serving as a strong feature that could lead the model to a trivial solution.

To examine this potentially trivial model behavior, we modified the dataset to include node 32 during training while mapping events with node 32 as normal. We randomly selected 500 events from the training set with the source node 31 and created 500 additional identical events, replacing the normal source node 31 with our suspicious node 32. These 500 modified events were then injected into the training data. If this alteration does not significantly reduce performance or produce significantly different results, while simultaneously mapping injected events closely to the enclosed ball centre in the training phase, our hypothesis regarding the undesirable trivial model behavior would be refused.

\begin{figure}[b]
\centering
\includegraphics[width=0.48\textwidth]{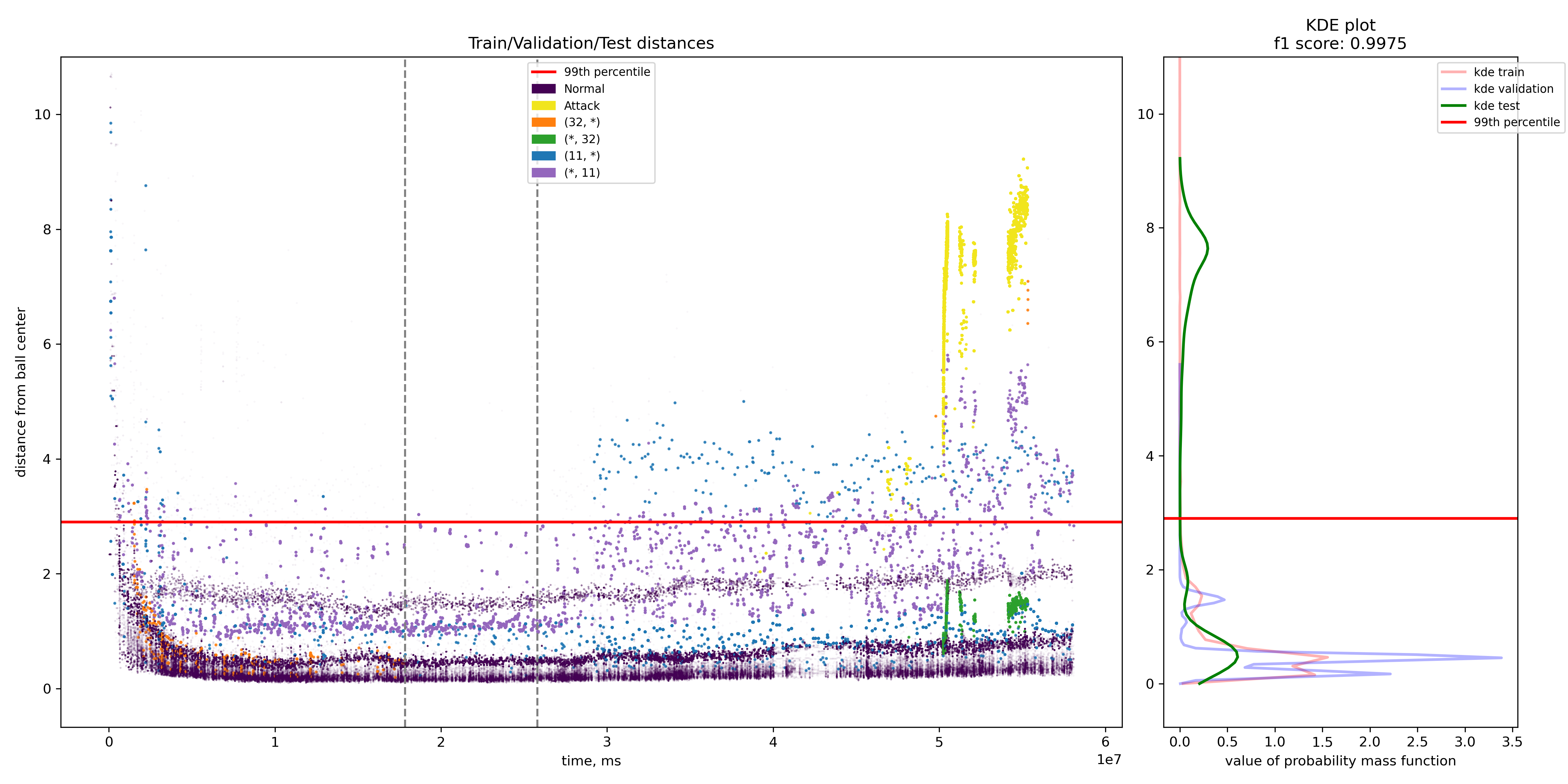}
\includegraphics[width=0.48\textwidth]{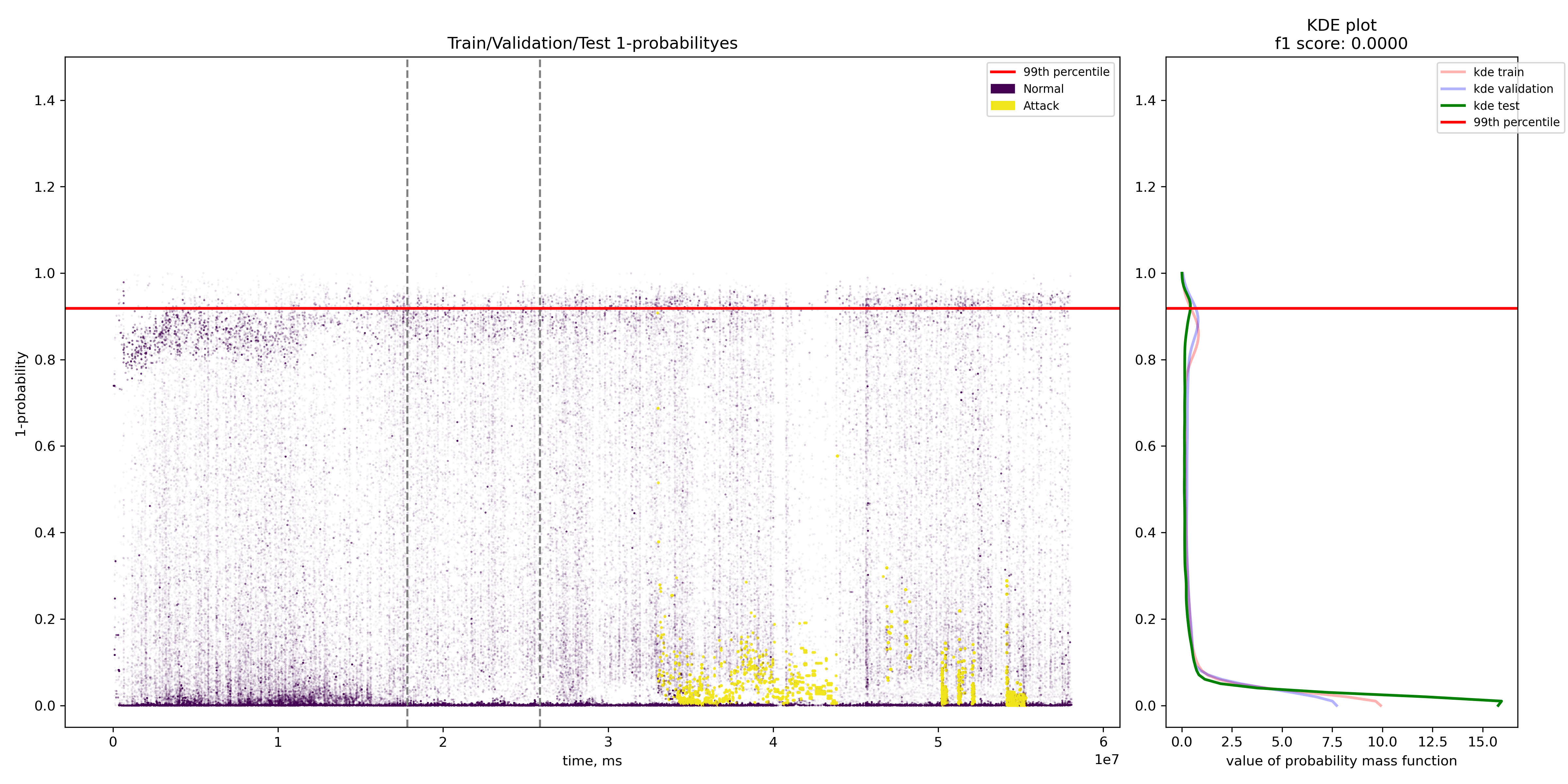}
\caption{Friday dataset. Left: TGN-SVDD with additional 500 events with the node 31 as a source injected in train data and labeled with orange. Right: Vanilla TGN with 1-p on the y-axis, where p is the probability of the event to occur.} \label{fig:fri_32}
\label{fig:fri_vanil_tgn}
\end{figure}

As illustrated in Figure \ref{fig:fri_vanil_tgn} (left), the model successfully learned to map injected events indistinguishably from the remaining normal activities, maintaining a good performance, as shown in Table \ref{tab:feat_32_metrics_fri}.

Similarly to TGN-SVDD we provide visualisation of vanilla TGN in the Figure \ref{fig:fri_vanil_tgn} (right). The results look noisy with many events assigned to a low probability. This does not allow to properly distinguish attack events from normal ones. One possible explanation is the strong assumption over negative sampling, which is made originally at random in the TGN paper and may lead to suboptimal solutions. More discussions about that can be found in the article \cite{poursafaei2022better}.

\section{Conclusion}

In this contribution, we presented a novel unsupervised model for intrusion detection, TGN-SVDD, making explicit use of the dynamic graph based behaviour of the data, by modelling network communications as a temporal dynamic graph.

For the evaluation, we pre-processed the public CIC-2017 dataset, which consists of 4 different attack days which we treated as different datasets and various modern attacks. We demonstrated that our method significantly outperforms classical techniques, as well as the vanilla TGN model. We demonstrated that our model can accurately identify the majority of specific attacks present in the datasets, while maintaining a moderate level of false positives.

In our experiments, we investigate potential limitations of the utilised dataset and suggest a possible remedy by including the attacker IP in the normal dataset, making the dataset more challenging. Our proposed model, however, still obtains high performance values as measured by our metrics. Future work includes the evaluation of data from other domains of anomaly detection. Also, investigating a semi-supervised approach such as the Deep semi-supervised SVDD, is a promising direction.

\begin{table}[t]
\centering
  \caption{The confusion matrix for all attack classes is presented, corresponding to the same experimental setup as previously described. To threshold the scores from TGN-SVDD, we selected the 99th percentile from the training set.}
  \label{tab:multiclass_results}
  \vspace{1mm}
  \begin{tabular}{ c @{\hspace{4mm}} l  c c  c c  c c}
\toprule
  & & \multicolumn{6}{c}{True Class}\\
  \addlinespace[2mm] 
  & & \multicolumn{6}{c}{\textit{Wednesday}}\\
  \cmidrule(l{.25em}){3-8}
  & &  Normal &  Golden Eye & Hulk & Heartbleed  & Slowloris & Slow http test\\ 
\cmidrule(l{.22em}){3-8}

\multirow{9}{*}{\rotatebox[origin=c]{90}{\makebox[0pt]{Predicted Class}}} & Normal  &   301788 &  0 &  0 & 0  &  0 & 0 \\
& Attack & 1764 & 2996 & 1 & 4217 & 3898 & 12538\\

  \cmidrule(l{.25em}){3-8}
  & & \multicolumn{4}{c}{\textit{Friday}} & \multicolumn{2}{c}{\textit{Tuesday}}\\
  \cmidrule(l{.25em}){3-6} \cmidrule(l{.25em}){7-8}
  & &  Normal &  Bot & DDoS & PortScan  & Normal & ssh/ftp-Patator\\ 
\cmidrule(l{.25em}){3-6} \cmidrule(l{.25em}){7-8}

& Normal  &   276697 & 1247 & 0 & 0 &  293213 & 0 \\
& Attack & 1471 & 0 & 44927 & 159126 & 1920 & 6954\\

 \cmidrule(l{.25em}){3-8}
  & & \multicolumn{6}{c}{\textit{Thursday}}\\
  \cmidrule(l{.25em}){3-7}
  & &  Normal & Brute Force & XSS & SQL Injection  & Infiltration & \\ 
\cmidrule(l{.22em}){3-7}

& Normal  &   287108 &  0 & 0 & 0  &  6 &  \\
& Attack & 55184 & 1365 & 661 & 12 & 0 & \\

\bottomrule
\end{tabular}
\end{table}

\begin{table}[t]
\centering
  \caption{Friday working hours. With additional 500 events with the node 31 as a source in train data. In the table established TGN-SVDD performance applying simple 99 percentile from train set on test set.}
  \label{tab:feat_32_metrics_fri}
  \setlength{\tabcolsep}{5pt}
  \begin{tabular}{ l c c c c }
\toprule
  \textit{Friday}
  &  Precision &    Recall &  F1-score &  ROC AUC \% \\ 
\midrule

TGN-SVDD         &   0.995 & 0.999 & 0.997 & 0.999 \\

\bottomrule
\end{tabular}
\end{table}

%
%

\bibliographystyle{splncs04}
\bibliography{bibliography}
\end{document}